\documentclass[runningheads]{llncs}
\usepackage{graphicx}
\usepackage{tikz}
\usepackage{comment}
\usepackage{amsmath,amssymb} % define this before the line numbering.
\usepackage{color}

\usepackage[colorlinks=true]{hyperref}
\usepackage[capitalise]{cleveref}
\usepackage{graphicx}
\usepackage{booktabs}
\usepackage{amsfonts}       % blackboard math symbols
\usepackage{nicefrac}       % compact symbols for 1/2, etc.
\usepackage{microtype}      % microtypography
\usepackage{xcolor}         % colors
\usepackage{bm}
\usepackage{multirow}
\usepackage{tabularx}
\usepackage{makecell}
\usepackage{ulem}

\newcommand{\etal}{\textit{et al}. }
\newcommand{\ie}{\textit{i}.\textit{e}., }
\newcommand{\eg}{\textit{e}.\textit{g}. }

\usepackage[accsupp]{axessibility}  % Improves PDF readability for those with disabilities.

\begin{document}
% \linenumbers
\pagestyle{headings}
\mainmatter
\def\ECCVSubNumber{5795}  % Insert your submission number here

\title{The Overlooked Classifier in \\ Human-Object Interaction Recognition} % Replace with your title

\begin{comment}
\titlerunning{ECCV-22 submission ID \ECCVSubNumber} 
\authorrunning{ECCV-22 submission ID \ECCVSubNumber} 
\author{Anonymous ECCV submission}
\institute{Paper ID \ECCVSubNumber}
\end{comment}
%******************

%\begin{comment}
\titlerunning{The Overlooked Classifier in HOI Recognition}

\begin{comment}
\author{
    Ying Jin\textsuperscript{\dag\ddag}~~
    Yinpeng Chen\textsuperscript{\dag}~~
    Lijuan Wang\textsuperscript{\dag}~~
    Jianfeng Wang\textsuperscript{\dag}~~\\
    Pei Yu\textsuperscript{\dag}~~
    Lin Liang\textsuperscript{\dag}~~
    Jenq-Neng Hwang\textsuperscript{\ddag}~~
    Zicheng Liu\textsuperscript{\dag}
    \\\\
    Microsoft\textsuperscript{\dag},~~~~University of Washington\textsuperscript{\ddag}\\
    {\tt\small\{ying.jin,yiche,lijuanw,jianfw,peyu,lliang,zliu\}@microsoft.com}\\{\tt\small\{jinying,hwang\}@uw.edu}
}
\end{comment}
\author{
    Ying Jin\textsuperscript{\dag\ddag} \and
    Yinpeng Chen\textsuperscript{\dag} \and
    Lijuan Wang\textsuperscript{\dag} \and
    Jianfeng Wang\textsuperscript{\dag} \and
    Pei Yu\textsuperscript{\dag} \and
    Lin Liang\textsuperscript{\dag} \and
    Jenq-Neng Hwang\textsuperscript{\ddag} \and
    Zicheng Liu\textsuperscript{\dag}
}
\authorrunning{Y. Jin et al.}
\institute{Microsoft\textsuperscript{\dag} \and University of Washington\textsuperscript{\ddag}}

%\email{\{ying.jin,yiche,lijuanw,jianfw,peyu,lliang,zliu\}@microsoft.com \{jinying,hwang\}@uw.edu}

%\end{comment}
%******************
\maketitle

\begin{abstract}
Human-Object Interaction (HOI) recognition is challenging due to two factors: (1) significant imbalance across classes and (2) requiring multiple labels per image. This paper shows that these two challenges can be effectively addressed by improving the classifier with the backbone architecture untouched. Firstly, we encode the semantic correlation among classes into the classification head by initializing the weights with language embeddings of HOIs. As a result, the performance is boosted significantly, especially for the few-shot subset. Secondly, we propose a new loss named LSE-Sign to enhance multi-label learning on a long-tailed dataset. Our simple yet effective method enables detection-free HOI classification, outperforming the state-of-the-arts that require object detection and human pose by a clear margin. Moreover, we transfer the classification model to instance-level HOI detection by connecting it with an off-the-shelf object detector. We achieve state-of-the-art without additional fine-tuning. %Code and models are available at \url{github.com/eccv22/defr.git}

\end{abstract}

\iffalse
\section{Introduction}
\begin{figure}
  \centering
  \includegraphics[width=0.5\linewidth]{figs/fig1_mAP.pdf}
  \caption{\textbf{\small Detection-Assisted v.s Detection-Free} HOI classifiers. Although the detection-free baseline (using ImageNet pre-trained ViT-B/32 \cite{vit20} as backbone) degrades severely from the detection-assisted HAKE \cite{hake19}, DEFR, our method, achieves significant improvement by leveraging language embedding as classifier initialization and the proposed LSE-Sign loss. DEFR/16 (ViT-B/16 backbone) achieves 65.6 mAP on the HICO dataset}
  \label{fig:1}
\end{figure}
\fi

\section{Introduction}
Human-Object Interaction (HOI) recognition has drawn significant interest for its essential role in scene understanding. HOI recognition aims to retrieve multiple \textit{$<$verb, object$>$} pairs (\eg, ``hold, apple'') in the image. Image-level HOI classification is challenging for two reasons. First, the datasets \cite{hico15,vg17} are long-tailed and have severe imbalance across classes. Second, it has multiple labels per image and the positions of the HOIs are unknown. Most of the recent work focuses on leveraging the object and keypoint detection while overlooking the classifier. We discover that the classifier is critical for both HOI classification and detection.

A crucial distinction between HOI recognition and conventional image classification is that the HOI classes are combinations of verbs and objects, so there is a stronger semantic correlation between the classes. For example, the average number of classes sharing the same verb is 5.1 in the HICO \cite{hico15} dataset, and the average number of classes sharing the same object category is 7.5. However, the effective use of such semantic cues is still underexplored in existing methods. Some work \cite{liao2020ppdm,uniondet20,sgg-imp17,atl21,scg20} classify the objects and verbs independently. This idea considerably reduces the number of classes to be predicted but is prone to polysemy because the same verb can have different meanings in different HOIs (\eg, ``{\em make}, pizza'', ``{\em make}, vase''). As one step forward, we use language models to generate text embeddings of HOI classes and use them to initialize the weight vectors in the linear classification layer. This approach, later referred to as \textit{Language Embedding Initialization}, offers superior performance compared with the widely used Random Initialization methods \cite{xavier10,kaiminginit15}, particularly in the few-shot cases.

\begin{figure}[t]
  \centering
  \includegraphics[width=0.8\linewidth]{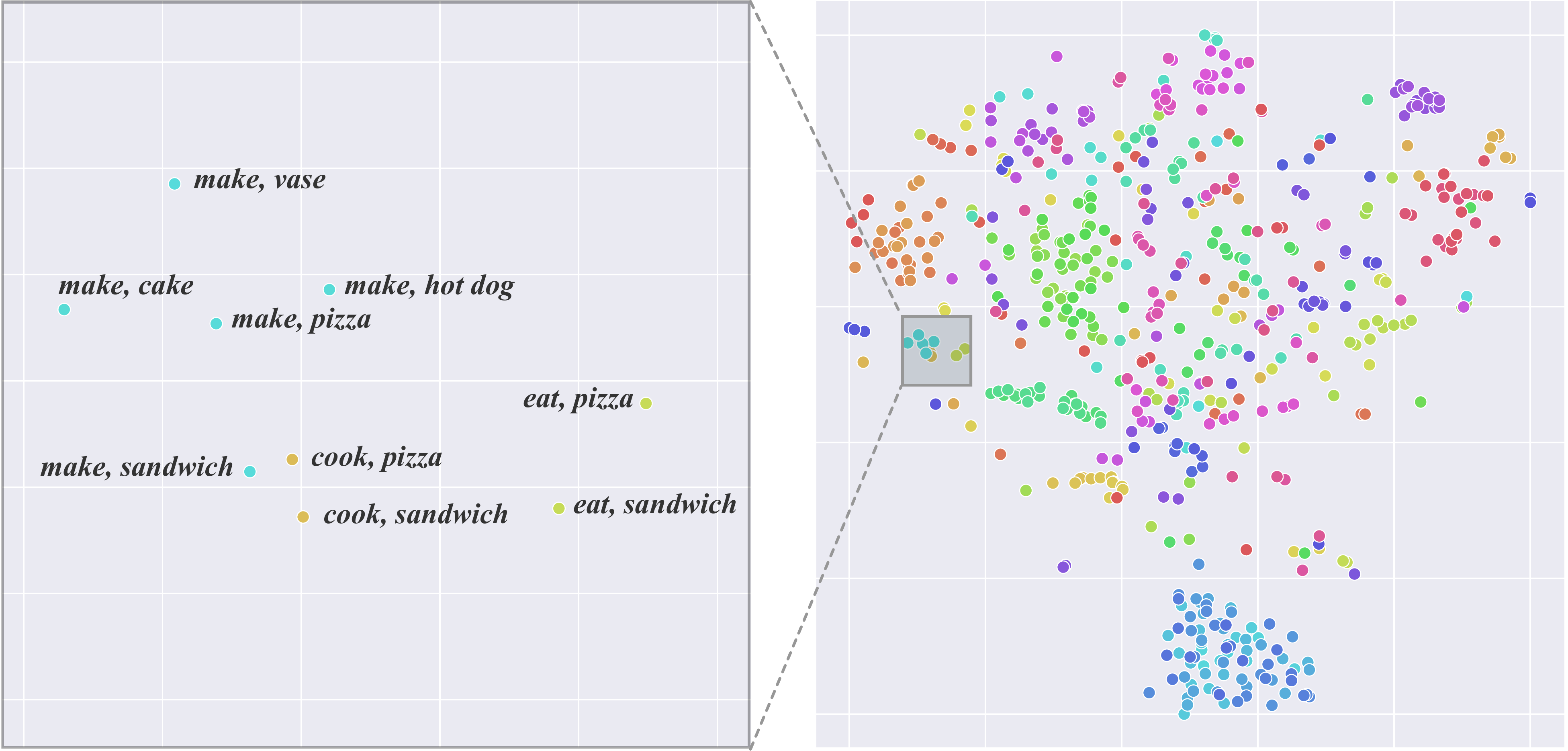}
  \caption{{\textbf{t-SNE visualization of BERT-encoded HOI embeddings}. Each point represents an HOI class. \textbf{Left:} zoomed view of a cluster. HOIs with closer semantic meanings (\eg, \textit{$<$make, pizza$>$, $<$cook, pizza$>$}) locate closely. \textbf{Right:} 600 HOI classes from the HICO dataset \cite{hico15} colored by the verb. The figure shows a clear clustering pattern, which captures the semantic structure. We use these embeddings as weight initialization}}
  \label{fig:2_bert}
\end{figure}

The rationale lies in the fact that the weight vectors in the linear classifier can be viewed as the proxy (feature center) of each class \cite{circleloss20,normface17}. When using language embeddings as weight initialization, the initial positions of proxies in the vector space are no longer random but are determined by their semantic structure. This configuration eases the multi-label training process as the vision feature from the backbone is guided to move toward the semantic feature center, where HOIs that frequently co-exist in an image also closely distribute (see \cref{fig:2_bert}). Furthermore, language embeddings provide proper weight initialization for few-shot classes, of which the weight vectors in the classifier cannot be properly learned with just a few samples. 

Existing studies \cite{transfer16,pairwise18} use {\it weighted} sigmoid cross-entropy loss \cite{wbce15} to handle the positive/negative imbalance per class. This is superior to cross-entropy loss but is less effective when the class is extremely imbalanced (20.1\% of the classes in HICO has $\leq$ 5 positive samples). We propose a new loss function named Log-Sum-Exp Sign (\textit{LSE-Sign}) loss to enhance such a multi-label learning scenario on long-tailed datasets. We design the loss function so that its gradient form is a softmax function of losses from all classes. The softmax function amplifies the loss from poorly classified classes and suppresses the loss from well-classified classes. This idea facilitates multi-label learning and balances the loss over all classes dynamically. Please note that the LSE-Sign loss is different from focal loss \cite{lin2017focal} in two ways: 1). it considers the {\em relative} difficulty among all classes while focal loss operates on each class, and 2) focal loss requires tuning two hyper-parameters.

Though a detection-free baseline using ImageNet-1K pre-trained ViT-B/32 as the backbone experiences a severe performance degradation compared with the detection-assisted state-of-the-arts (from 47.1 to 37.8 mAP), Language Embedding Initialization using BERT successfully boosts it to 50.6 mAP. The applicability generalizes to other types of language models, including SimCSE \cite{simcse21} and CLIP \cite{clip21}. LSE-Sign loss provides an average of 3.1 mAP improvement in eight experiments (\cref{tab:bce_vs_sign}). Based upon these two findings, we outperform existing detection-assisted approaches \cite{pairwise18,pastanet20,hake19} by a clear margin. With a ResNet101 backbone, our method achieves 53.6 mAP, surpassing the state-of-the-art \cite{hake19} that uses both object and human keypoint detections by 6.5 mAP. With a ViT-B/16 backbone, our method achieves 65.6 mAP. On few-shot subsets, we achieve 52.7 mAP on 1-shot and 56.9 mAP on 5-shot classes. 

On the instance level, we decouple the HOI detection task to localization and regional HOI classification. The advantage of being detection-free allows us to apply the classification model on a regional human-object box pair produced by any off-the-shelf detector. With a vision transformer backbone, we convert bounding boxes into an attention mask so that the CLS token can only attend to the region of interest. Without additional training under instance-level supervision, it achieves state-of-the-art (SOTA) performance (32.35 mAP) on the HICO-DET dataset, outperforming the SOTA \cite{cdn21} which is trained on HICO-DET.

\begin{figure*}[t]
  \centering
  \includegraphics[width=0.6\linewidth]{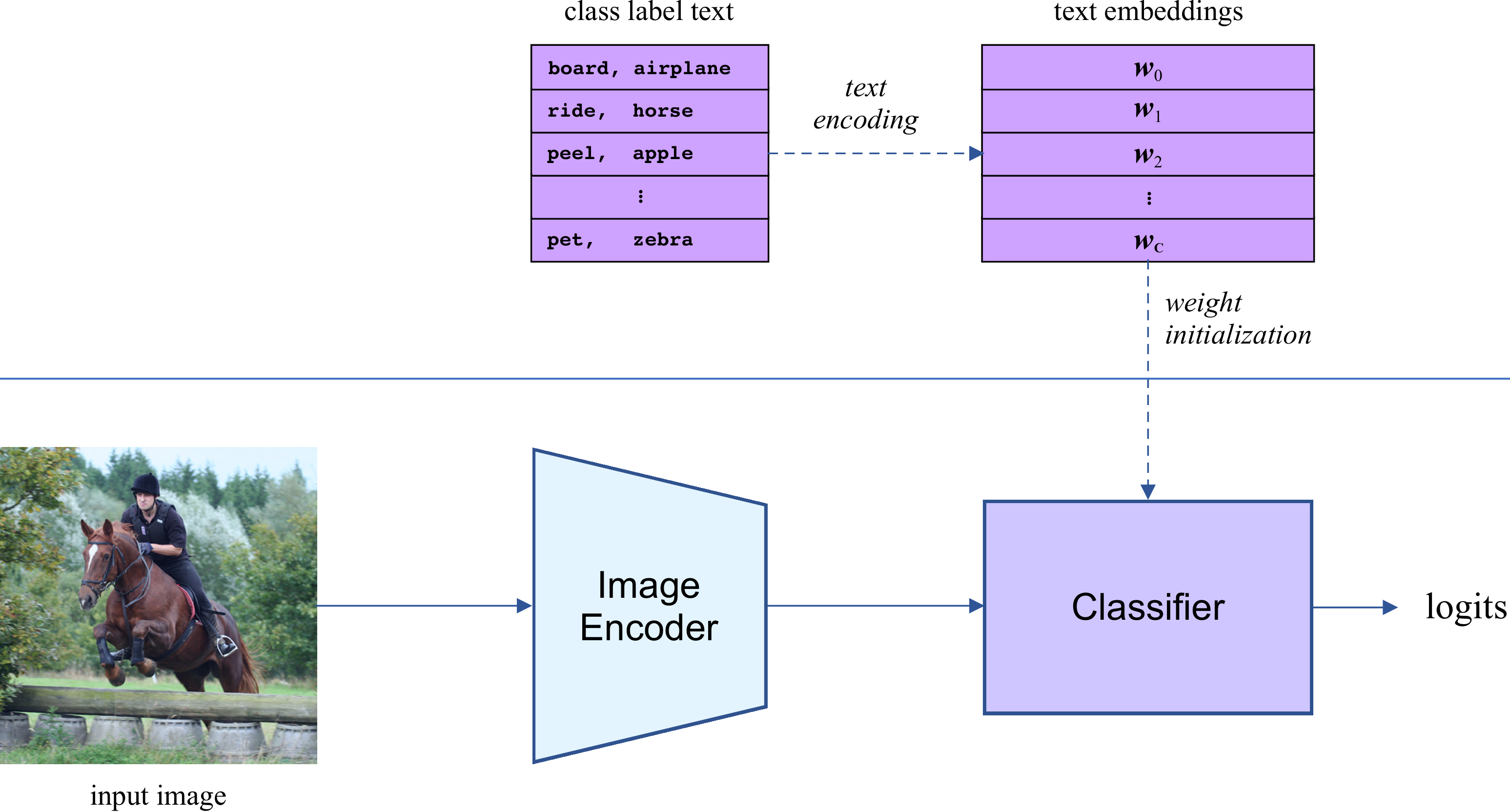}
  \caption{\textbf{DEtection FRee (DEFR) HOI recognition pipeline}. It has an image encoder and a linear classifier. The weight of the linear classifier is initialized by $w$, the text embeddings of HOI classes. We call this \textit{language embedding initialization}, detailed in \cref{sec:embedding_init}. Compared to detection-assisted approaches, DEFR significantly simplifies the pipeline}
  \label{fig:pipeline}
\end{figure*}

\section{Related work}

\noindent\textbf{HOI Classification:}
This is an image-level multi-label classification task, where HOI labels are given without specifying the positions of the human and object. Existing work \cite{rcnn15,transfer16,pairwise18,pastanet20} successfully leverage object detectors to locate humans and objects in the first stage and then infer interactions between human-object pairs. Though the performance gets improved over time, the pipelines are becoming more and more complicated. Since the HOI labels are not grounded to specific human-object pairs, Multiple Instance Learning (MIL) \cite{mil98} is required in these work to enable training. Girdhar \etal \cite{attentionalpooling17} removes MIL by utilizing human pose guided attention maps. PaStaNet \cite{pastanet20} and HAKE \cite{hake19} achieve performance boost by utilizing PaSta, which are action classes on the body part level (\eg, ``right\_hand: hold\_something''), and require additional annotations to train. In contrast, this paper shifts focus to a simplified \textit{detection-free} solution. Unlike existing work, our perspective is to leverage the semantic correlations among HOI classes and design a suitable loss function. The proposed method emphasizes the overlooked classification head and complements previous studies.

\vspace{0.4em}\noindent\textbf{HOI Detection:}
The detection task requires a pair of human-object boxes and the corresponding HOI classes, and instance-level supervision is provided. Existing work can be categorised into three streams. Two-stage methods \cite{ican18,kaiming_hoi18,drg20,fcm20,pastanet20,kim2020detecting,scg20} require object detection prior to HOI classification to extract regional features. Graph neural networks are often used \cite{scg20,sgg-imp17} to classify the verbs between human-object pairs. For training, two-stage methods usually initialize their backbone from a pre-trained object detector. One-stage methods \cite{liao2020ppdm,uniondet20,ggnet21,atl21,ipnet20} mostly execute object detection and HOI detection in parallel and match them afterwards. Recent studies \cite{asnet21,hotr21,e2e21,qpic21,cdn21} achieve end-to-end HOI detection based on DETR \cite{detr2020} and benefit from the wider perception field of transformers \cite{qpic21}.

\iffalse
\vspace{0.4em}\noindent\textbf{Scene Graph Generation:}
As in connection with HOI, scene graphs \cite{vg17} are also pairwise relationships. A scene graph is a graph representation in which the nodes are objects and edges are relationships. The IMP \cite{sgg-imp17} method detects objects first and then region features are iteratively fine-tuned in a graphical network. To reduce the compute, \cite{sgg-factorizable18} constructs sparsely connected sub-graphs, and \cite{sgg-grcnn18} applies a Relationship Proposal Network. \cite{sgg-neuralmotif18} reveals the strong statistical bias in the Visual Genome \cite{vg17} dataset, and \cite{sgg-unbiased20} suggests unbiased post-processing to neutralize the model from biased training. \cite{sgg-knowledge-embed19} and \cite{sgg-commonsense20} use frequency priors from the dataset and text as external knowledge to improve performance.
\fi

\vspace{0.4em}\noindent\textbf{Language Models:} Natural language is occasionally used in related work as statistical priors or additional features. DRG \cite{drg20} and HAKE \cite{hake19} use BERT-generated embeddings of object class or part-level actions (PaSta) as features. BERT \cite{bert18} is trained in an unsupervised manner by predicting masked tokens in the sequence (\ie masked language modeling). The BERT embedding space is not isotropic, which could harm the performance in downstream tasks. SimCSE \cite{simcse21} and other methods \cite{sbert19,bertflow20,bertwhite21} improve the sentence embedding of BERT. CLIP \cite{clip21} and ALIGN \cite{align21} jointly train a language model and a vision encoder.

\section{Method}
In this section, we introduce the implementation of Language Embedding Initialization and the LSE-Sign loss. The two ideas complement the classifier while being agnostic to the backbone architecture. Our DEFR method has a simple pipeline (shown in \cref{fig:pipeline}), including a vision transformer \cite{vit20} as the backbone and a linear layer as the classifier. The classifier is initialized with language embeddings instead of the conventional random initialization methods \cite{xavier10,kaiminginit15}. 

\subsection{Language Embedding Initialization}
\label{sec:embedding_init}

HOI classes have strong semantic correlations as they are combinations of verbs and objects. Language models can effectively encode the structure of HOI classes, which can be hard for the image backbone to learn along (see \cref{fig:2_bert}). Based on this observation, we use language embedding to initialize the linear classification layer. By doing so, the semantic cues are encoded into the classifier, which guides training and provides proper weight initialization for few-shot classes in the classifier. This is referred to as \textit{Language Embedding initialization}.

Specifically, we first convert the HOI classes (\eg, ``$<ride, bicycle>$'') to prompts (\eg, ``\verb+a person riding a bicycle+'') and then generate language embeddings with a language model. The embeddings are normalized to be consistent in the scale, and then used as the initial weight in the linear classification layer. The bias of the linear classifier is zero-initialized. During training, the output logit for the $i^{th}$ class is the dot product of the image feature and $w_i$, a row vector in the classifier's weight $w$, plus bias (see \cref{fig:pipeline}). Since dot product is the unnormalized cosine similarity, $w_i$ is often considered as the proxy for a given class \cite{circleloss20,normface17}. 

\begin{figure*}[t]
  \centering
  \includegraphics[width=1\linewidth]{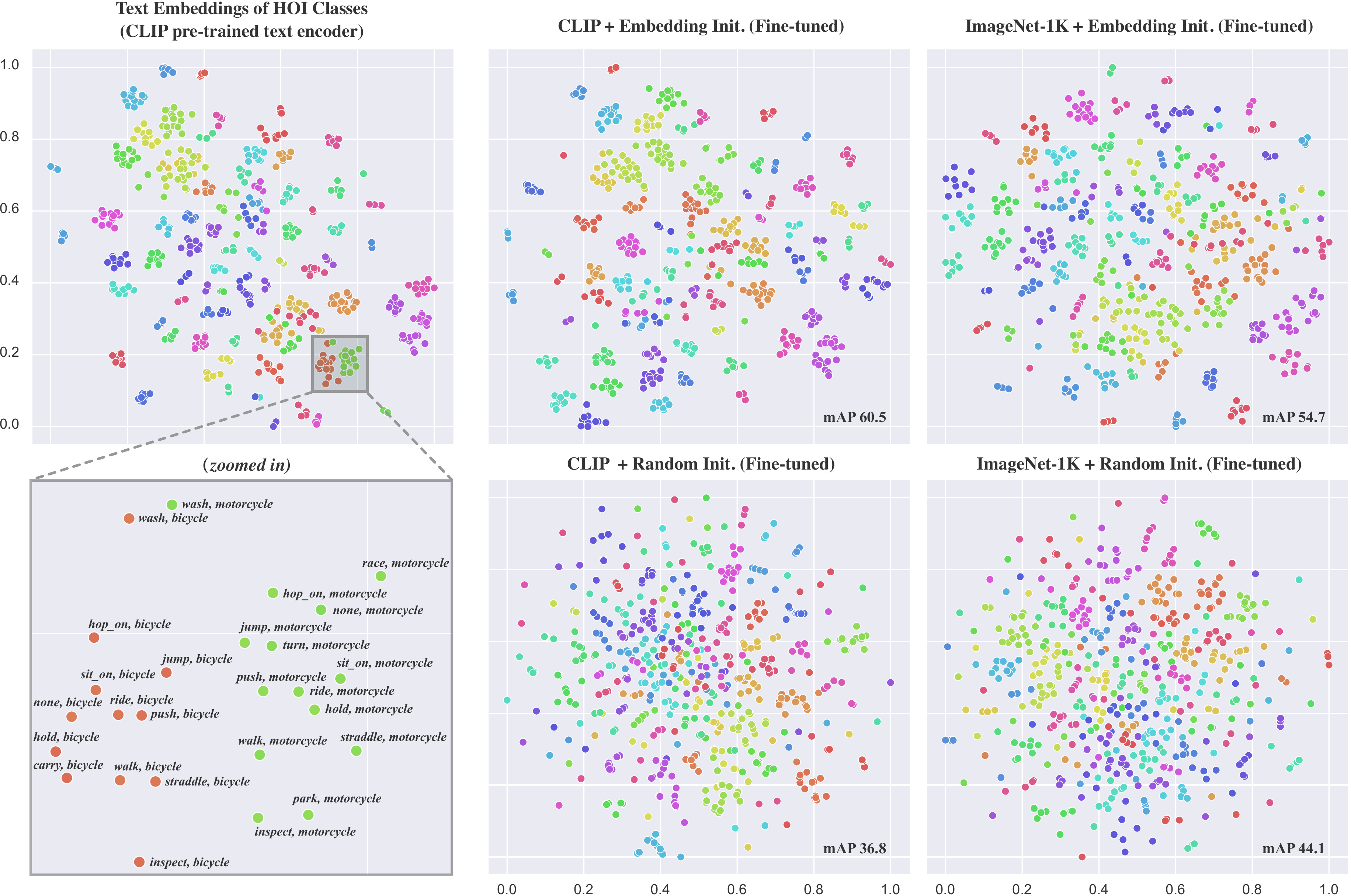}
  \caption{\textbf{t-SNE visualization of classifier weights.} \textbf{Left Column:} same as \cref{fig:2_bert} but the language model is CLIP. \textbf{Middle Column:} the weight vectors (proxies) of 600 HOI classes after fine-tuning, when CLIP pre-trained ViT-B/32 backbone is used. The top and bottom use language embedding initialization and random initialization, respectively. \textbf{Right Column:} the weight vectors (class proxies) after fine-tuning, when ImageNet-1K pre-trained ViT-B/32 backbone is used. The top and bottom use text embedding and random initialization for the linear classifier, respectively. Text embeddings are clearly clustered before fine-tuning, and the overall structure changes slightly after fine-tuning. However, this clustered structure is not clearly learned when fine-tuned with random initialization, reflected in lower model performance (see mAP results in the plots)}
  \label{fig:t-SNE-class-embedding}
\end{figure*}

In our scenario, language embedding initialization using BERT fires a performance boost of 12.8 mAP for an ImageNet-1K pre-trained ViT-B/32 backbone. More importantly, the applicability is agnostic to backbone architectures (ViT and ResNet \cite{resnet16}), pre-training, and the language model being used. The gain is maximized when the language model is jointly trained with the image encoder (\eg, CLIP). \cref{fig:t-SNE-class-embedding} visualizes the linear classifier's weight vectors per class before (left-column) and after fine-tuning with both CLIP (middle-column) and ImageNet-1K (right-column) pre-trained backbones. The weight vectors initialized with language embeddings maintain clustered after fine-tuning, meaning the initialization is closer to optimal. However, this structure is difficult to learn when fine-tuned with the random initialization, resulting in lower performance (shown in \cref{fig:t-SNE-class-embedding} bottom row).

\subsection{Log-Sum-Exp Sign (LSE-Sign) Loss}
We propose a log-sum-exp sign (LSE-Sign) loss function to facilitate multi-class learning, as an image usually contains multiple interactions (\eg, \textit{cut carrot, hold carrot, peel carrot}). Let $\bm{x}$ denote a feature vector from the backbone and $\bm{y}=\{y_1, y_2,\dots,y_C\}$ denote multiple class labels, where $C$ is the number of HOI classes, and $y_i \in \{1, -1\}$, indicating the positive and negative labels, respectively. We first pass the $\gamma$ scaled vision feature to the last linear layer as:
\begin{align}
  \label{eq:si}
  \;\;\;\;\;\;\;\;\;s_i = \gamma \bm{x}^T \bm{w}_i + \bm{b}_i \;\;\;\;\;\;\;\;\;\;\;\;\;\;\;\;\;\;\;\;\;\;\;\;\;
\end{align}
where $\bm{w}_i$ is the $i^{th}$ row in the weight matrix in the linear classifier, corresponding to the $i^{th}$ HOI class. It is initialized with a normalized text embedding. $\gamma$ is a scalar hyper-parameter controlling the range (the effect is ablated in \cref{tab:loss_gamma}). The losses of both positive and negative classes can be unified into $e^{-y_is_i}$, where the label $y_i$ controls the sign. The overall loss is defined as the log-sum-exp function of $-y_is_i$ as follows:
\begin{align}
    \label{eq-loss}
      \;\;\;\;\;\;\;\;\;\mathcal{L} &= \log\left(1+\sum_{i=1}^C e^{-y_is_i}  \right) \;\;\;\;\;\;\;\;\;\;\;\;
                   % & = \log\left(1+\sum_{i\in pos} e^{-s_i} + \sum_{i\in neg} e^{s_i} \right),
\end{align}
where the constant term 1 in the log function sets the zero lower bound for the loss. Log-sum-exp is a smooth approximation of the maximum function, and its gradient is the softmax function as follows:

\begin{align}
  \label{eq-gradient}
    \;\;\;\;\;\;\;\;\frac{\partial \mathcal{L}}{\partial s_i} &= \frac{-y_ie^{-y_is_i}}{1+\sum_{j=1}^C e^{-y_js_j}} \\
    &=\frac{-1_{i \in pos}e^{-s_i}+1_{i \in neg}e^{s_i}}{1+\sum_{i\in pos} e^{-s_i} + \sum_{i\in neg} e^{s_i}} %\log\left(\sum_{i=1}^C e^{-y_is_i} \right).
\end{align}
where $1_{i \in pos}$ is a Dirac delta function that returns $1$ if the $i^{th}$ class is positive and $0$ otherwise. Compared to the binary cross entropy loss and focal loss that consider each class separately, LSE-Sign loss considers the dependency across classes as the magnitude of the gradients are normalized over all classes and distributed by a softmax function. This facilitates multi-label learning on an long-tailed dataset as it encourages learning of classes with larger loss values and suppresses the learning of classes with smaller loss values.

\subsection{HOI Detection}
We discover that a strong HOI classification model trained on image-level labels can aid HOI detection effectively. We connect our frozen classification model with an off-the-shelf object detector to recognize the regional HOI. The two models are trained separately and when combined as a whole, they achieve state-of-the-art performance \textit{without} additional fine-tuning on instance-level supervision.

To connect the classifier with the detector, we convert the detected bounding boxes to self-attention masks, $\Phi$. The last transformer layer in the ViT backbone consumes $\Phi$ so that the \verb+CLS+ token attends only to the specified region of interest:

\begin{equation}
    \small
    \label{eq:mask}
    Attention(Q, K, V) = softmax(\Phi+\frac{QK^T}{\sqrt{d_k}}) V
\end{equation}
\Cref{eq:mask} is the \textit{Attention} \cite{transfer16} function of a transformer layer. $\Phi$ is a binary mask converted from the bounding boxes. $\Phi_{i, j}$ equals $-\infty$ if $i$ is the CLS token and $j$ a patch outside the given bounding boxes, and 0 otherwise. $d_k$ is the dimension of $Q, K$ and $V$.

\begin{figure}[h]
  \centering
  \includegraphics[width=0.9\linewidth]{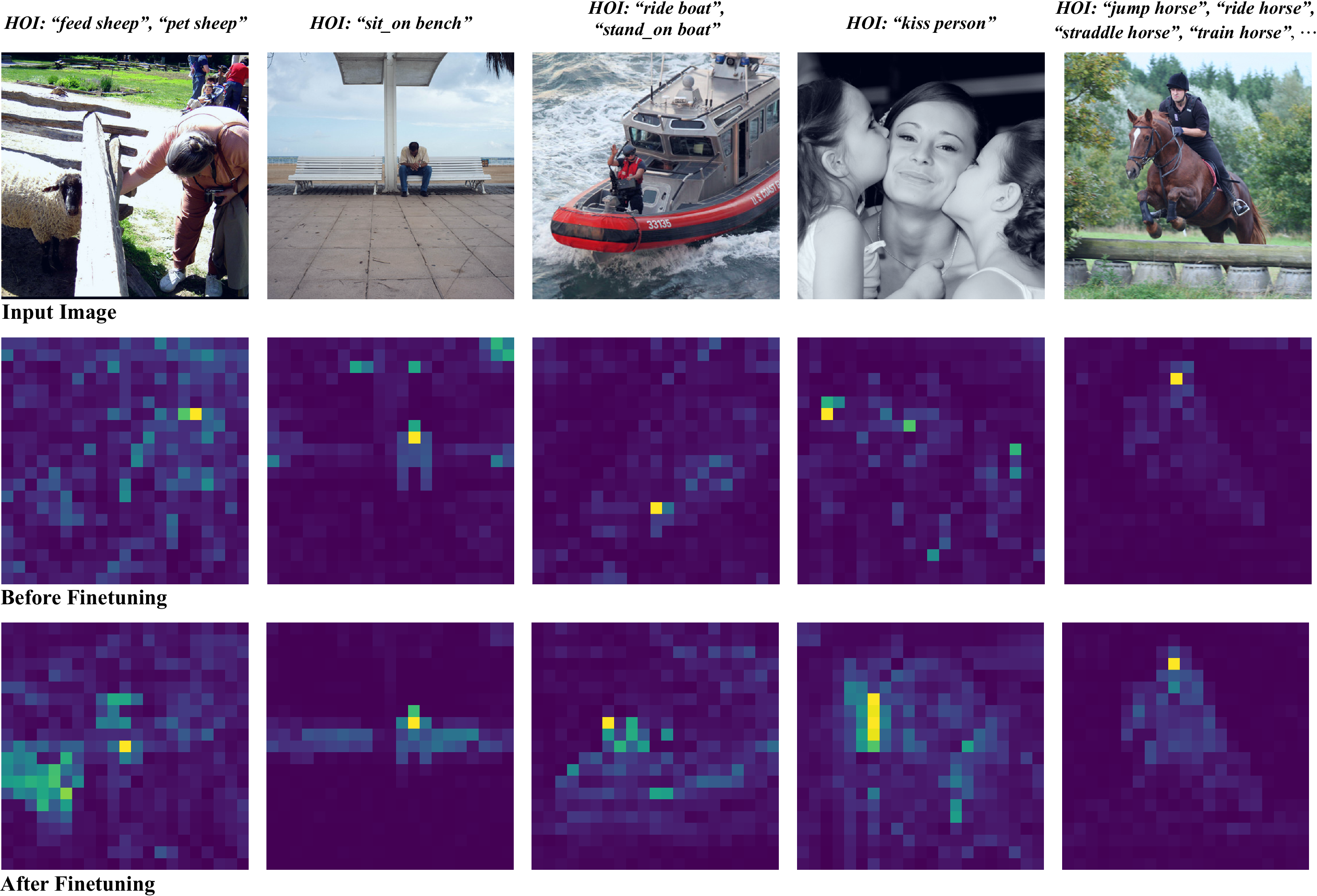}
  \caption{\textbf{Visualization of CLS attention map}. First row: the input image. Second row: the attention map of the CLIP pre-trained backbone without fine-tuning on HICO. Third row: the attention map of DEFR fine-tuned on HICO. After fine-tuning, the attention activates more on the HOI related objects and activates less elsewhere. The feature for HOI learned in the classification task aids HOI detection effectively}
  \label{fig:vis-cls-attn}
\end{figure}

Therefore, the CLS token in the last layer attends only to the region of interest specified by the pair of human and object bounding boxes. Object probabilities are multiplied to corresponding HOI classes. By doing so, we treat HOI detection as a special case of HOI classification, as in this task, DEFR classifies only the regional HOI. This pipeline requires no training on instance-level HOI annotation thanks to the properly learned feature in the classification task (see \cref{fig:vis-cls-attn}).

\subsection{Relationships with Existing Work}
Existing work on image-level HOI classification \cite{rcnn15,transfer16,pairwise18,pastanet20,sgg-imp17} successfully leverage object detectors and human pose to improve the feature for HOI classification. In contrast, we emphasizes on improving the classification head, which is under-explored in the field. Our method achieves a simplified detection-free pipeline and is complementary to existing studies.

On HOI detection, we tackle the problem by reusing the classification model on the instance-level. As the detector is trained separately, our method is easy to leverage future advancements on object detection. Our approach decouples object detection from HOI recognition, which is different from existing work that contains a detector as a module in the network.

\section{Experiments}

We evaluate the proposed DEFR on both HOI classification and HOI detection. 
\subsection{Implementation for HOI Classification}
\noindent\textbf{Datasets:}
For image-level HOI classification, we conduct experiments on two commonly used datasets: HICO and MPII. The HICO dataset \cite{hico15} contains 600 HOI categories which have 117 unique verbs and 80 COCO \cite{coco14} object classes. Each image may contain multiple HOI classes and multiple human-object pairs. The training set has $38,116$ images and test set has $9,658$ images. Following existing methods, we randomly reserve 10\% images from the training set for validation and report performance on the test set. The MPII dataset \cite{mpii14} contains 15,205 training images and 5,708 test images. Unlike HICO, each image is labeled with only one of 393 interaction classes. We follow \cite{transfer16,pairwise18} to report performance on the validation set that contains 6,987 images.

\vspace{0.4em}\noindent\textbf{Pre-training:} The backbone is ViT-B/32, and we investigate three different strategies to pre-train the backbone at resolution 224.
\begin{enumerate}
  \item Image classification task on ImageNet-1K or ImageNet-21K referred to as CLS1K and CLS21K, respectively.
  \item Masked language modeling task motivated by~\cite{kim2021vilt}, where the network input includes both an image and the associated text. Both modalities are fully attended to each other in each transformer block. The pre-training datasets are Google Conceptual Captions~\cite{SoricutDSG18}, SBU~\cite{OrdonezKB11}, COCO~\cite{coco14} and Visual Genome~\cite{vg17}. This is referred to as MLM.
  \item Image–text contrastive learning task based on CLIP~\cite{clip21}. The image encoder is jointly trained with a text encoder and the two modalities are contrasted on the encoder output representations. Only the image encoder is used as the backbone. Here, we directly use the released CLIP model\footnote{\url{https://github.com/openai/CLIP}}, and reuse the term CLIP to refer to the image encoder as pre-training.
\end{enumerate}

\noindent\textbf{Fine-tuning:} All models have a ViT-B/32 backbone fine-tuned at resolution 672 with the AdamW \cite{adam14} optimizer without weight decay. We use a batch size of 128 on 8 V100 GPUs. We set the base learning rate as 1.5e-5 for CLIP pre-trained backbones and 1e-4 otherwise, and use cosine scheduling with warm restarts \cite{sgdr16} every 5 epochs. The best model is fine-tuned for 10 epochs. Data augmentation of random color jittering, horizontal flipping, and resized cropping is used. To reduce class imbalance, we adopt over-sampling so that each class has at least 40 samples per epoch.

\begin{table}[t]
  \caption{\textbf{The gain of individual components} evaluated on HICO. The baseline uses ImageNet-1K pre-trained ViT-B/32 as the backbone, random initialization for the classifier, and binary cross entropy loss. Language embeddings generated by BERT are used as classifier's weight initialization}
  \label{tab:path}
  \centering
  \small
  \setlength{\tabcolsep}{1.6mm}{
  \begin{tabular}{llccc}
    \toprule
    \textbf{} & \textbf{Pre-training} & \textbf{\makecell{LSE-Sign\\Loss}} & \textbf{\makecell{Embedding\\Initialization}} & \textbf{mAP}  \\ \midrule
    Baseline  & CLS1K             &                          &                        & 37.8          \\ \midrule
              & CLS1K             & \checkmark               &                        & 44.1          \\
              & CLS1K             &                          & ~\checkmark~BERT      & 50.6          \\
              & CLS1K             & \checkmark               & ~\checkmark~BERT      & 53.5          \\
    %          & CLS1K             & \checkmark               & \checkmark~CLIP      & 54.7          \\
    %DEFR/32   & CLIP              & \checkmark               & \checkmark~CLIP      & \textbf{60.5} \\
    %DEFR/16   & CLIP              & \checkmark               & \checkmark~CLIP      & \textbf{65.6} \\
    \bottomrule
  \end{tabular}
  }
\end{table}
% It is worth mentioning that using the CLIP image backbone along scores 34.2 mAP after fine-tuned on this task, much lower than our final solution. 
\vspace{0.4em}\noindent\textbf{Improved Classifier:} \cref{tab:path} shows the path from baseline to DEFR. The baseline achieves 37.8 mAP, much lower than the detection-assisted approaches. The LSE-Sign loss gains 6.3 mAP, and language embedding initialization adds another 9.4 mAP with BERT as the language model. We found the performance is further improved when the language model is jointly trained with the image backbone (\ie using CLIP, to 60.5 mAP). This demonstrates that our proposed approaches are effective and complementary mechanisms for HOI recognition.

\subsection{Comparision on HOI Classification}
\noindent\textbf{HICO dataset}: \cref{tab:mAP} compares our detection-free method (DEFR) with the prior works that need assistance from object detection or human keypoint detection. Our method achieves significantly better accuracy without detection assist. Specifically, DEFR achieves 65.6 mAP on the HICO, gaining over the state-of-the-art HAKE \cite{hake19} by 18.5 mAP.

\vspace{0.4em}\noindent\textbf{Few-shot analysis}: Our model outperforms existing methods considerably in few-shot subsets on HICO, by leveraging language embedding initialization and LSE-Sign loss. As  shown in \cref{tab:fewshot}, DEFR achieves 52.7 mAP in one-shot classes, only 20\% lower than the all-class performance.

\begin{table}[t]
  \centering
  \small
  \caption{\textbf{Comparison with the state of the art on HICO}. Dependencies (additional input) required by existing methods include \emph{Bbox}: object detection, \emph{Pose}: human keypoints, \emph{PaSta} \cite{pastanet20}: additional training data of part level actions. We report the performance of DEFR/32 (ViT-B/32 backbone) and DEFR/16 (ViT-B/16 backbone)}
  \label{tab:mAP}
\setlength{\tabcolsep}{3.2mm}{
  \begin{tabular}{ l c c c c }\toprule
    \multirow{2}{*}{\textbf Method}                             & \multicolumn{3}{c }{\textbf{Dependencies}} & \multirow{2}{*}{\textbf mAP}                                     \\
    \cmidrule(lr){2-4}
                                                        & \textbf{Bbox}                             & \textbf{Pose}        & \textbf{PaSta} &                 \\
    \cmidrule(lr){1-1}
    \cmidrule(lr){2-4}
    \cmidrule(ll){5-5}
    R*CNN~\cite{rcnn15}                                 & \checkmark                                &                      &                 & $28.5$          \\
    Girdhar~\textit{et al.}~\cite{attentionalpooling17} &                                           & \checkmark           &                 & $34.6$          \\
    Mallya~\textit{et al.}~\cite{transfer16}            & \checkmark                                &                      &                 & $36.1$          \\
    Pairwise-Part~\cite{pairwise18}                          & \checkmark                                & \checkmark           &                 & $39.9$          \\
    PastaNet~\cite{pastanet20}                          & \checkmark                                & \checkmark           & \checkmark      & $46.3$          \\
    HAKE~\cite{hake19}                          & \checkmark                                & \checkmark           & \checkmark      & $47.1$          \\
    \midrule
    ViT-B/32 Baseline (CLS1K) & & & & $37.8$ \\
    ViT-B/32 Baseline (CLIP) & & & & $34.2$ \\
    \textbf{DEFR/32}                                         &                                           &                      &                 & $\textbf{60.5}$ \\
    \textbf{DEFR/16}                                         &                                           &                      &                 & $\textbf{65.6}$ \\
    \bottomrule
  \end{tabular}
  }
\end{table}

\begin{table}[t]
  \caption{\textbf{Few-shot performance} evaluated on HICO. Few@i means classes that the number of training images is $i$. The number of HOI classes for Few@1, 5, 10 are 49, 125 and 162, respectively. Methods can be different in the backbone and detection assists}
  \label{tab:fewshot}
  \centering
  \small
  \setlength{\tabcolsep}{2.3mm}{
  \begin{tabular}{lcccc}
    \toprule
    \textbf{Method}     & \textbf{mAP} & \textbf{Few@1} & \textbf{Few@5} & \textbf{Few@10}  \\ \midrule
    Pairwise-Part~\cite{pairwise18}  &    39.9   &   13.0   &   19.8  & 22.3 \\ 
    PastaNet~\cite{pastanet20}       &    46.3   &   24.7   &   31.8  & 33.1 \\
    HAKE~\cite{hake19}               &    47.1   & 25.4     &   32.5  & 33.7 \\ \midrule
    \textbf{DEFR/16}     &    \textbf{65.6}   & \textbf{52.7}    &   \textbf{56.9}  & \textbf{57.2}\\
    \bottomrule
  \end{tabular}
  }
\end{table}

\vspace{0.4em}\noindent\textbf{MPII dataset}: we evaluate HOI classification on the MPII dataset in addition to HICO. As shown in \cref{tab:mpii}, our model achieves 43.6 mAP with the same ResNet101 backbone without detection assist. This proves that our proposed approach is effective on ResNet architectures as well.

\begin{table}[t]
  \caption{\textbf{Comparison on the MPII dataset}. We additionally apply the proposed approach on ResNet101 backbone for fair comparision. We follow \cite{rcnn15,transfer16,pairwise18} to report performance on the validation set that contains 6,987 images. Different from HICO, the MPII dataset has only one interaction label per image}
  \label{tab:mpii}
  \centering
  \small
  \setlength{\tabcolsep}{6.5mm}{
  \begin{tabular}{llc}
    \toprule
    \textbf{Method}   & \textbf{Backbone} & \textbf{mAP}\\ \midrule
    R*CNN~\cite{rcnn15}  & VGG16 & 21.7 \\
    Girdhar~\textit{et al.}~\cite{attentionalpooling17} & ResNet101 &  30.6 \\ 
    Pairwise-Part~\cite{pairwise18}      &   ResNet101   &   32.0 \\\midrule
    \textbf{DEFR}     &   ResNet101  & \textbf{43.6}\\
    \textbf{DEFR}     &   ViT-B/16  & \textbf{55.3}\\
    \bottomrule
  \end{tabular}
  }
\end{table}

\subsection{Ablations}
Several ablations were performed that focus on key components of DEFR: (a) the image backbone, (b) language embedding initialization and (c) the loss function.

\vspace{0.4em}\noindent\textbf{Pre-training:}
We evaluate three different pre-training tasks for the backbone: (a) image classification (CLS1K/CLS21K), (b) masked language modeling (MLM), and (c) image-text contrastive learning (CLIP). \cref{tab:pre-train_init} (the first column) shows the results for these pre-training tasks using random initialization in the linear classifier.

\vspace{0.4em}\noindent\textbf{Classifier initialization:}
\cref{tab:pre-train_init} compares the conventional random initialization and language embedding initialization. Clearly, the language embedding initialization with both language models provides a consistent improvement over all pre-training methods. It is worth noting that ImageNet pre-trained models (CLS1K/CLS21K) gain 10+ points from language embedding initialization.

\begin{table}[t]
  \centering
  \small
  \caption{\textbf{Ablations on backbone pre-training and classifier initialization}. Numbers are mAP on the HICO dataset. \textit{Random}: the default random initialization; \textit{Embedding}: language embedding initialization using BERT or CLIP's text encoder. The backbone is ViT-B/32 and fine-tuned with LSE-Sign loss}
  \label{tab:pre-train_init}
  \setlength{\tabcolsep}{2mm}{
  \begin{tabular}{l c c c}
    \toprule
    \multirow{2}{*}{\textbf{Pre-training}} & \multicolumn{3}{c}{\textbf{Classifier Initialization}}                        \\
    \cmidrule(lr){2-4}
                     & \makecell{\textbf{Random}\\\textbf{Initialization}}   & \makecell{\textbf{BERT}\\\textbf{Embedding}}             & \makecell{\textbf{CLIP}\\\textbf{Embedding}}      \\
    \midrule
    CLS1K                        & $44.1$            & $53.5_{(+9.4)}$            & $54.7_{(+10.6)}$      \\
    CLS21K                       & $44.2$            & $53.9_{(+9.7)}$            & $55.1_{(+10.9)}$      \\
    MLM                          & $43.6$            & $47.0_{(+3.4)}$            & $47.1_{(+3.5)}~~$      \\
    CLIP                         & $36.8$            & ~$51.0_{(+14.2)}$      & $\bm{60.5}_{(+23.7)}$ \\
    \bottomrule
  \end{tabular}
  }
\end{table}

\begin{table}[t]
  \centering
  \small
  \caption{\textbf{Binary Cross Entropy (BCE) Loss vs. LSE-Sign Loss} evaluated on HICO for four differently pre-trained models. The classifier is initialized with CLIP text embeddings}
  \label{tab:bce_vs_sign}
  \setlength{\tabcolsep}{5.9mm}{
  \begin{tabular}{l c c}
    \toprule
    \multirow{2}{*}{Pre-training} & \multicolumn{2}{c}{Loss Function}                        \\
    \cmidrule(lr){2-3}
                                 & BCE Loss       & LSE-Sign Loss      \\
    \midrule
    &\multicolumn{2}{c}{\small BERT Embedding Initialization~~~~~} \\
    CLS1K                        & $50.6$       & $53.5_{(+2.9)}$      \\
    CLS21K                       & $51.0$       & $\bm{53.9}_{(+2.9)}$      \\
    MLM                          & $45.8$       & $47.0_{(+1.2)}$      \\
    CLIP                         & $44.4$       & $51.0_{(+6.6)}$ \\
    \midrule
    &\multicolumn{2}{c}{\small CLIP Embedding Initialization~~~~~} \\
    CLS1K                        & $51.5$       & $54.7_{(+3.2)}$      \\
    CLS21K                       & $50.0$       & $55.1_{(+5.1)}$      \\
    MLM                          & $46.6$       & $47.1_{(+0.5)}$      \\
    CLIP                         & $57.9$       & $\bm{60.5}_{(+2.6)}$ \\
    \bottomrule
  \end{tabular}
  }
\end{table}

\vspace{0.4em}\noindent\textbf{Loss function:}
The choice of loss function is vital in fine-tuning. Previous works use binary cross entropy (BCE) loss and treat HOI recognition as a set of binary classification problems. \cref{tab:bce_vs_sign} shows that the proposed LSE-Sign loss outperforms BCE loss on all four differently pre-trained backbones. \cref{tab:clip_loss} compares LSE-Sign loss with other alternatives: binary cross entropy (BCE) loss, weighted BCE loss and focal loss~\cite{lin2017focal} on our model.
Weighted cross entropy loss is intended to impose a weight on the loss of each class so that each category is balanced among the positive samples and negative samples. It is not effective if the dataset is severely long-tailed.
Focal loss~\cite{lin2017focal} reduces the weight of the massive negative samples, but requires manual tuning of two hyper-parameters. Our LSE-Sign loss outperforms all others by a clear margin.\\

\noindent\textbf{Scalar $\gamma$ in \cref{eq:si}:} LSE-Sign loss has a scalar $\gamma$ which controls the magnitude of output per class $s_i$. It is needed as the initial weights $w_i$ are normalized embeddings. $\gamma$ performs like the temperature parameter of softmax. A small $\gamma$ makes softmax distribution close to uniform and a large $\gamma$ makes the softmax close to one-hot. The best trade-off is achieved when $\gamma=100$, see \cref{tab:loss_gamma}.

\begin{table}[t]
\small
  \caption{\textbf{Comparison between LSE-Sign loss and other loss functions} evaluated on HICO. We change the loss function of our best model with ViT-B/32 backbone and language embedding initialization. Weighted BCE is the binary cross entropy loss weighted by positive-negative-ratio per-class. Focal loss uses $\gamma$=2 and $\alpha$=0.25 as recommended in \cite{lin2017focal}}
  \label{tab:clip_loss}
  \centering
  \setlength{\tabcolsep}{12.0mm}{
  \begin{tabular}{lc}
    \toprule
    {\textbf{Loss function}} & {\textbf{mAP}}      \\
    \midrule
    Weighted BCE           & $54.7$        \\
    BCE                    & $57.9$        \\
    Focal Loss             & $53.2$        \\
    \midrule
    LSE-Sign Loss (ours)   & \textbf{60.5} \\
    \bottomrule
  \end{tabular}
  }
\end{table}

\begin{table}[t]
    \small
  \caption{\textbf{Ablation of scalar} $\gamma$ in \cref{eq:si} evaluated on HICO dataset. The highest accuracy is achieved at $\gamma=100$. The backbone is CLIP pre-trained ViT-B/32, the classifier is embedding initialized and LSE-Sign loss is used for fine-tuning}
  \label{tab:loss_gamma}
  \centering
  \setlength{\tabcolsep}{4.0mm}{
  \begin{tabular}{lccccc}
    \toprule
    $\bm{\gamma}$ & 50   & \textbf{100} & 150  & 300  & 500  \\
    \midrule
    \textbf{mAP}      & 60.4 & \textbf{60.5} & 59.1 & 57.2 & 53.0 \\
    \bottomrule
  \end{tabular}
  }
\end{table}

\vspace{0.4em}\noindent\textbf{Backbone architecture:}
We train DEFR with different backbone architectures on HICO as in \cref{tab:backbone}. We see all backbones outperform the current SotA \cite{hake19} by using our proposed methods.

\begin{table}[t]
    \small
  \caption{\textbf{Ablation of backbone architecture} on HICO dataset. We apply our method on different ResNet and ViT backbone architectures. The pre-training of the backbone, language embedding initialization and loss function stay the same as our best model. DEFR method with ResNet-50 achieves 49.7 mAP, still outperforms the SotA \cite{hake19} by 2.6 mAP}
  \label{tab:backbone}
  \centering
  \setlength{\tabcolsep}{12.0mm}{
  \begin{tabular}{lc}
  \toprule
    {\textbf{DEFR Backbone}} & {\textbf{mAP}}      \\
    \midrule
    ResNet-50           & $49.7$        \\
    ResNet-101          & $53.6$        \\
    ViT-B/32             & $60.5$        \\
    ViT-B/16             & $65.6$        \\
    \bottomrule
  \end{tabular}
  }
\end{table}

\subsection{Comparison on HOI Detection}
\vspace{0.4em}\noindent\textbf{Dataset}: We evaluate HOI detection performance on the HICO-DET dataset \cite{hicodet18}. HICO-DET contains 117,871 annotated human--object pairs in the training set and 33,405 in the test set. HICO-DET and HICO share the same images and classes, but HICO-DET provides bounding box localized HOI annotations.

\vspace{0.4em}\noindent\textbf{Evaluation metric}: Following existing studies, we use the standard evaluation protocol \cite{hicodet18}. Each positive prediction should have the correct HOI class together with a pair of human-object bounding boxes with IoU greater than 0.5 in reference to ground truth. Similar to the classification task, the average precision (AP) of each HOI class is separately computed and then averaged (mAP).

\vspace{0.4em}\noindent\textbf{Results}: \cref{tab:results_det} compares the HOI detection performance on the HOI-DET dataset. Detectors used in this table are fine-tuned on the HICO-DET dataset. We use the detector from \cite{drg20,scg20}. 

Though DEFR is never trained under instance-level HOI supervision from HICO-DET, with stronger backbones, our method outperforms existing SotA that are trained on HICO-DET. Specifically, on rare sets (classes with less than 10 samples), our method outperforms current studies by a clear margin, which is consistent with our image-level classification results. Different from previous works, our method showcases the strong correlation between HOI-Classification and HOI-Detection in a very simple format that HOI-Detection is a special case of HOI-Classification with additional input of bounding boxes of human and objects. Note that the detected boxes are obtained from an object detector trained separately.

\begin{table}[t]\small
	\caption{\textbf{HOI Detection performance} on HICO-DET~\cite{hicodet18}. \underline{Full}: full set of 600 HOI classes, \underline{Rare}: a subset of 138 HOI classes that have less than 10 training samples, \underline{Non-rare}: the rest 462 classes. We use the same detector as \cite{drg20,scg20}. Our method is not fine-tuned on HICO-DET, but with stronger backbones, we outperform existing work trained on HICO-DET}
	\label{tab:results_det}
	\setlength{\tabcolsep}{3mm}
	\begin{tabularx}{\linewidth}{l l c c c}
		\toprule
		\textbf{Method} & \textbf{Backbone}  & \textbf{Full} & \textbf{Rare} & \textbf{Non-rare} \\
		\midrule
		\multicolumn{5}{c}{\textit{Two-stage and One-stage Methods}}\\
		\midrule
		PPDM~\cite{liao2020ppdm} & Hourglass-104 & 21.94 & 13.97 & 24.32 \\
		Bansal \etal~\cite{bansal2020} & ResNet-101 & 21.96 & 16.43 & 23.63 \\
		GG-Net~\cite{ggnet21} & Hourglass-104 & 23.47 & 16.48 & 25.60 \\
		VCL~\cite{hou2020} & ResNet-50 & 23.63 & 17.21 & 25.55 \\
		DRG~\cite{drg20} & ResNet-50-FPN & 24.53 & 19.47 & 26.04 \\
		IDN~\cite{li2020} & ResNet-50 & 26.29 & 22.61 & 27.39 \\
		ATL~\cite{atl21} & ResNet-50 & 28.53 & 21.64 & 30.59 \\
		SCG~\cite{scg20} & ResNet-50-FPN & \textbf{31.33} & \textbf{24.72} & \textbf{33.31} \\
		\midrule
		\multicolumn{5}{c}{\textit{DETR-based End-to-End Methods}}\\
		\midrule
		HOI-Trans~\cite{e2e21} & ResNet-50 & 23.46 & 16.91 & 25.41 \\
		QPIC~\cite{qpic21} & ResNet-50 & 29.07 & 21.85 & 31.23 \\
        HOTR~\cite{hotr21} & ResNet-50 & 25.10 & 17.34 & 27.42 \\
        AS-Net~\cite{asnet21} & ResNet-50 & 28.87 & 24.25 & 30.25 \\
        CDN-B~\cite{cdn21} & ResNet-50 & 31.78 & 27.55 & 33.05 \\ 
        CDN-L~\cite{cdn21} & ResNet-101 & \textbf{32.07} & \textbf{27.19} & \textbf{33.53} \\ 
        \midrule
		\textbf{Ours}~ & ViT-B/16 & 32.35 & 33.45 & 32.02 \\
		\textbf{}~ & ViT-L/14 & \textbf{35.36} & \textbf{38.28} & \textbf{34.49} \\
		\bottomrule
	\end{tabularx}
\end{table}

\iffalse
\section{Discussion of Limitations}
Although this paper reveals that language embedding initialization and LSE-Sign loss function play important roles for detection-free HOI recognition, we have not shown if these two components can work together with detection supervision to further push the state of the arts. In HOI detection, our performance on non-rare classes is lower than the state of the art. We will investigate these in future work.
\fi

\section{Conclusion}
In this paper, we discover that classifier is critical for both HOI classification and detection, though overlooked in existing studies. Our method builds upon two findings. Firstly, we show that the structure of HOI classes can be effectively leveraged by using language embedding as classifier initialization. Secondly, we propose the LSE-Sign loss to facilitate multi-label learning on a long-tailed dataset. A combination of DEFR and an object detector achieves the state-of-the-art HOI detection without additional fine-tuning. Our proposed detection-free method simplifies the pipeline and achieves higher accuracy than the detection-assisted counterparts. We hope that our work opens up a new direction for HOI recognition.

\clearpage
\bibliographystyle{splncs04}
\bibliography{refs}
\end{document}